\documentclass[runningheads]{llncs}
\usepackage{graphicx}
\usepackage{subcaption}
\usepackage{float}

\usepackage{lineno,hyperref}
\modulolinenumbers[5]
\usepackage[linesnumbered,boxed]{algorithm2e}
\usepackage{algpseudocode}
\usepackage{multirow}

\usepackage{environ}         
\usepackage{etoolbox}        
\begin{document}
\title{Cross-Lingual Query-by-Example Spoken Term Detection: A Transformer-Based Approach}
%
%
\author{Fatemeh Allahdadi \and
	Rahil Mahdian Toroghi \and
	Hassan Zareian
}
\institute{Iran Broadcasting University (IRIBU), Tehran, Iran 	
	\\
	\email{\{fatima.iribu@gmail.com\}},	\email{\{mahdiantr1974@ieee.org\}},
	\email{\{zareian@iribu.ac.ir\}}
}
\maketitle              
\vspace{-4mm}
\begin{abstract}
Query-by-example spoken term detection (QbE-STD) is typically constrained by transcribed data scarcity and language specificity. This paper introduces a novel, language-agnostic QbE-STD model leveraging image processing techniques and transformer architecture. By employing a pre-trained XLSR-53 network for feature extraction and a Hough transform for detection, our model effectively searches for user-defined spoken terms within any audio file. Experimental results across four languages demonstrate significant performance gains (19-54$\%$) over a CNN-based baseline. While processing time is improved compared to DTW, accuracy remains inferior. Notably, our model offers the advantage of accurately counting query term repetitions within the target audio.
	
%

\end{abstract}
\section{Introduction}\vspace{-2mm}
Query-by-Example Spoken Term Detection (QbE-STD) is a long-term speech retrieval task aimed at identifying all temporal instances within an audio corpus where a user-provided spoken query occurs \cite{8}. This task relies on direct acoustic matching, differentiating it from keyword spotting (KWS) which employs textual queries \cite{11,13,15,16}. While KWS is constrained to predefined terms, QbE-STD offers the flexibility to search for arbitrary spoken phrases \cite{12}.

%
\begin{figure}[!ht]
	\centering
	\includegraphics[width=12.3cm, height=6.1cm]{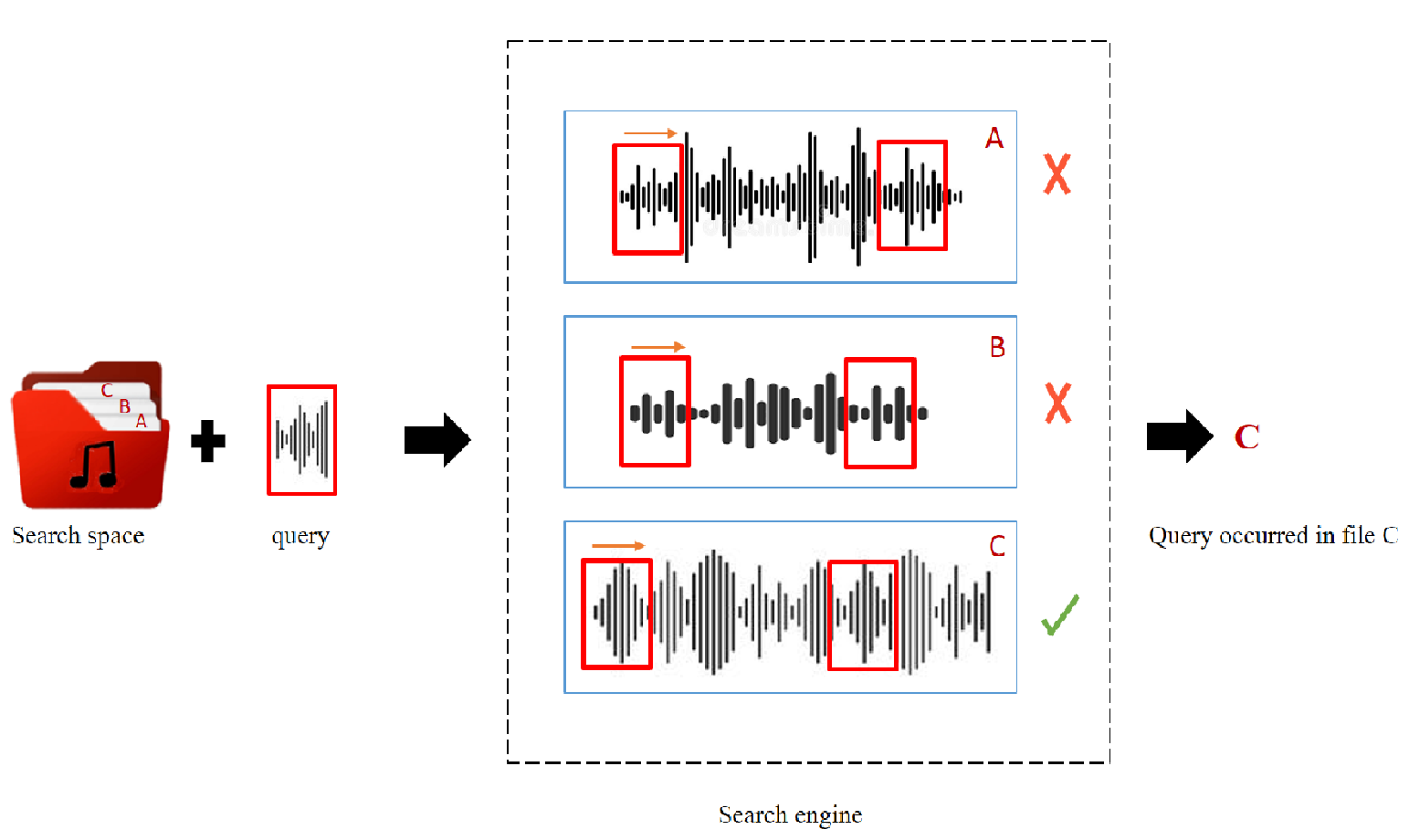}
	\caption{\small  Overview of the Query-by-Example Spoken Term Detection (QbE-STD)}\vspace{-6mm}
	\label{fi1}
\end{figure}

Early QbE-STD approaches relied on a two-stage pipeline: automatic speech recognition (ASR) to transcribe audio data followed by text-based information retrieval \cite{18,19,20}. While effective for high-resource languages with extensive transcribed data, these methods suffer from limitations in low-resource scenarios and are susceptible to out-of-vocabulary terms. Consequently, their applicability across languages is restricted \cite{36}.

%

In data-scarce environments where transcribed speech corpora are limited, a two-stage paradigm has emerged as a practical approach to query-by-example spoken term detection (QbE-STD). The initial stage involves extracting frame-level feature representations from audio signals. Traditionally, features such as Mel-Frequency Cepstral Coefficients (MFCCs) \cite{29,37} and posteriograms \cite{38} have been employed for this purpose. More recently, neural network-based feature extraction techniques have gained prominence  \cite{8,11,26,27}. The subsequent stage entails matching these extracted features against corresponding representations from spoken queries. Dynamic Time Warping (DTW) has been a commonly used algorithm for this alignment process \cite{8,11,21}. However, the performance of DTW-based systems is often hindered by their susceptibility to errors in frame-level feature representation and their computational inefficiency when dealing with large-scale audio datasets \cite{22}.

%
%


An alternative to DTW for query-by-example spoken term detection (QbE-STD) involves the combination of Acoustic Word Embeddings (AWEs) and Convolutional Neural Networks (CNNs). The core principle of AWE is to map variable-length speech segments into fixed-dimensional vector representations \cite{22,23,24,25}. This process entails training a Deep Neural Network (DNN) to generate embeddings such that semantically similar spoken terms are projected into proximity within the embedding space, while dissimilar terms are mapped to distant regions. This embedding learning is typically achieved through a weakly supervised training paradigm utilizing spoken term pairs.

 These approaches have demonstrated promising results in automatic speech recognition  \cite{30} and isolated word discrimination tasks \cite{31,32,33}. However, the inherent challenge of QbE speech search lies in the absence of explicit word boundaries within the target audio, rendering it a more complex problem than isolated word recognition  \cite{24}. To address this, recent research has explored the application of approximate nearest neighbor search techniques to AWE representations as a viable solution \cite{34,35}.


%

To conduct the search, a fixed-length window is incrementally shifted across the audio signal. However, this approach does not ensure comprehensive or consistent phonetic representation throughout the search space. By employing simple distance metrics such as cosine similarity, the computational overhead associated with Dynamic Time Warping (DTW) or complex neural network architectures like Recurrent Neural Networks (RNNs) \cite{22,23,25,40} and CNNs can be significantly reduced \cite{24}.

CNN-based approaches transform the spoken term detection (STD) problem into a binary classification task. By extracting features such as bottleneck features \cite{26,27,28} or posteriogram representations \cite{12,29} from speech signals and computing distance matrices between query and reference frames, image-like representations are generated. The presence of a query within the reference manifests as a quasi-diagonal pattern within the image (class one), while its absence is indicated by the lack of such a pattern (class zero). These image representations are subsequently employed to train a CNN classifier. However, this approach necessitates image resizing to a fixed dimension and careful class balancing, which can distort temporal information inherent in the audio signal. Additionally, CNN-based models typically exhibit a high demand for training data.
While existing methods have demonstrated theoretical promise, their practical applicability remains limited. This paper presents a novel, training-free approach to address the challenges posed by the requirement for transcriptions and segmentation. The key contributions of this work are:
\begin{itemize}
	\item A novel audio search method is proposed for efficient location of spoken terms within audio files.
	\item The system accommodates both microphone input and uploaded audio clips, providing file names and occurrence counts.
	\item A diverse, four-language dataset was constructed to evaluate the proposed method.
	\item A training-free model is introduced, demonstrating superior performance compared to existing approaches.
\end{itemize}
The remaining parts of this paper is structured as follows. In section \ref{se2}, we introduce the proposed approach in details. In section \ref{se3}, experimental evaluations, introduction of baseline models and, evaluation metrics are provided. Finally, the conclusion and future work are given in section \ref{se4}, preceded by the references being cited in this paper.
\vspace{-1mm}
\section{Methodology} 
\begin{figure}[!ht]
	\centering
	\includegraphics[width=11.9cm, height=4.5cm]{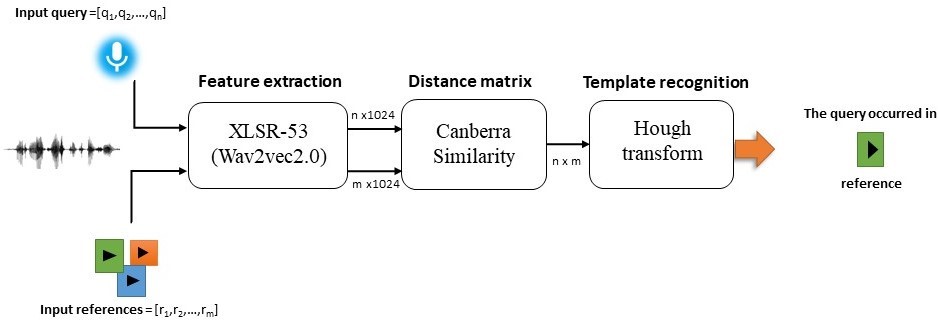}
	\caption{\small Block diagram of the proposed QbE-STD method}
	\vspace{-4mm}
	\label{fi2}
\end{figure}

\vspace{-3mm}
\label{se2}
A transformer-based QbE-STD model is proposed, that comprises three modules: feature extraction, distance matrix computation, and pattern recognition (Fig. \ref{fi2}). Novel approaches are introduced for feature extraction and distance matrix calculation. \vspace{-4mm}
\subsection{Feature Extraction}
Facebook AI's wav2vec 2.0 model introduced self-supervised learning for speech representation. This approach surpasses semi-supervised methods by learning from raw audio without transcriptions. The model's ability to capture underlying speech structures aligns more closely with human language acquisition.


The wav2vec 2.0 model employs a convolutional neural network to extract features from raw audio, followed by a masked language modeling approach using a transformer network. This self-supervised learning framework enables the model to learn robust speech representations with minimal labeled data. The transformer is trained to predict the original masked speech units, thereby learning robust representations of speech. Notably, the model requires only ten minutes of labeled data for fine-tuning.

\begin{figure}[!ht]
	\centering
	\includegraphics[width=13cm, height=5cm]{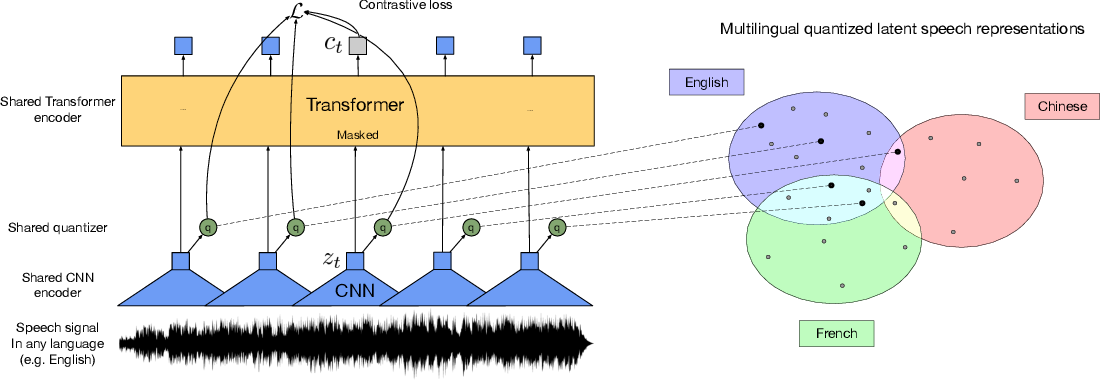}
	\caption{\small \textbf{The XLSR approach}. A shared quantization module over feature encoder representations
		produces multi-lingual quantized speech units whose embeddings are then used as targets for a
		transformer trained by contrastive learning. The model creatins bridges across languages \cite{2}.}
	\label{fi3}
\end{figure}

The XLSR-53 model, based on wav2vec 2.0, leverages cross-lingual pre-training on 53 languages and 56k hours of speech data\footnote{CommonVoice, BABEL, Multilingual LibriSpeech (MLS)} \cite{2}. It employs a convolutional encoder to extract 512-dimensional latent features from raw audio, which are then fed into both a quantizer for self-supervised learning and a 24-layer transformer encoder for context representation. The model is trained using a contrastive loss function. For the proposed QbE-STD system, the 11th block of the pre-trained XLSR-53 transformer is utilized for feature extraction. The details of the wav2vec2.0 model structure are shown in Fig.\ref{fi3}.
\vspace{-2mm}
%
%
%
%
\subsection{Distance Matrix}
\begin{figure}[!ht]
	\centering
	\subfloat[The query did not occur in the reference]{\includegraphics[width=7.6cm, height=2.2cm]{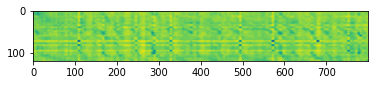}}
	\qquad
	\subfloat[The query ocurred in the reference ]{\includegraphics[width=7.6cm,height=2.2cm]{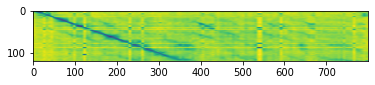}}
	\caption{\small Similarity matrix image in 0 and 1-class}
	\vspace{-7mm}
	\label{fi4}
\end{figure}

Frame-level comparison is performed by calculating a distance matrix between query and reference speech representations. These representations are extracted using the XLSR-53 model, resulting in $M \times 1024$ dimensional matrices (M is the number of frames in the speech file). The Canberra similarity metric is employed to compute pairwise distances between corresponding frames of the query and reference speech \cite{3}.
\vspace{-3mm}
\begin{equation}
\label{eq1}
d^{CAD}(q_i,r_j) = \sum_{k=0}^{l-1}
\frac{|{q_{i,k}-r_{i,k}}|}{|{q_{i,k}}|+|{r_{i,k}}|}
\end{equation}
where $ q_i $ is the i-th frame of query, $ r_j $ is the j-th frame of reference, and k is the dimension of a frame's feature vector (here 1024). If query has $ n $ frames and reference has $ m $, the obtained distance matrix is in $ n \times m $. The obtained matrix is saved as an image. If the query has occurred in the reference, the image will contain a quasi-diagonal line (as in Fig.\ref{fi4}(b)), otherwise, such a pattern will not be observed, see Fig.\ref{fi4}(a). The distance matrix is visualized as an image where the x-axis represents reference frames and the y-axis represents query frames.\vspace{-5mm}
\subsection{Template Recognition}
\vspace{-2mm}
In the template recognition phase, the paradigm shifts from speech signal analysis to image processing. The input for this stage consists of the images of distance matrix, with the objective of performing binary classification. To detect the diagonal patterns indicative of query occurrence within the reference, we employ a two-step approach. Initially, we apply the \textit{Canny edge detection} algorithm to identify salient edges within the distance matrix images. Subsequently, we utilize the Hough transform to isolate and characterize edges that correspond to the desired diagonal structures.
This methodology leverages the robustness of edge detection algorithms in conjunction with the parametric representation capabilities of the Hough transform. Such an approach allows for effective identification of linear structures within the distance matrices, which are crucial for determining query matches in the reference audio. The transition from temporal signal analysis to spatial pattern recognition in this stage represents a novel integration of diverse signal processing paradigms in the context of spoken term detection.

%
\indent
The Canny edge detector is an operator that uses a multi-step algorithm like Alg.\ref{al1} to find a wide range of edges in an image\cite{5}.\\\\
\scalebox{0.8}{%
\begin{algorithm}[H]
		\caption{Canny edge detection algorithm}
		\label{al1}
		\small	\bf A = Input image \\
		\textbf{\small Noise removal:} $ S = g(x,y) * A = \frac{1}{\sqrt{2\pi}\sigma}\exp^{\frac{x^2+y^2}{2\sigma^2}}$\\
		\textbf{\small Derivative:} \For{pixle(x,y)}{$\nabla S = \nabla (g*A) \to$ magnitude= $ \sqrt{S^2_x+S^2_y} $ ,
			direction = $ \theta = \arctan \frac{S_y}{S_x} $ 
		}
		\textbf{Non-maximal suppression:} \If{$ |\nabla S|(x_1,y_1) $ is largest gradient of edge }{Check neighbors in edge direction\\
			\If{$ |\nabla S|(x_1,y_1) \ge |\nabla S|(x_m,y_n) \forall m,n $}{$ M(x,y)= |\nabla S|(x_1,y_1) $
				\Else{$ M(x,y)=0 $}}}
		\textbf{Hysteresis Thresholding:} \For{every pixle (x,y)}{\If{$|\nabla S| > T_{upper}$ \textbf{OR} $|\nabla S| > T_{lower}$ \textbf{AND} next to an edge}{"is an edge pixel"
				\Else{"is a non-edge pixel"}}}
\end{algorithm}}
\vspace{0.4cm}\\
\indent
The identification of the reference file containing the target query necessitates the detection of diagonal patterns within the distance matrices. To accomplish this, we employ the Hough transform for straight line detection \cite{4}. The fundamental principle underlying the Hough transform is the transformation of a point grouping problem in Cartesian space into a peak detection problem in parameter space.
Leveraging this concept, a set of collinear points $(x, y)$ in the image plane can be represented in polar coordinates by the Eq.(\ref{eq2}):
\vspace{-2mm}
\begin{equation}
\label{eq2}
\rho = x\cos \theta + y\sin \theta
\end{equation}
\noindent
Where $ \theta $ is the angle of the line with respect to the origin, which varies between zero and $ \pi $, and $ \rho $ is the distance of the line to the origin. Pixels at the edge of the image, detected using the Canny edge detector, can provide sufficient evidence for the presence of a line in the image. These pixels fill the cells of an accumulator array with the parameters  $ \theta $ and $ \rho $, then the cells that show stronger lines in the image are selected by applying a threshold limit. The points that are on a line in the image intersect each other at a point in the parametric space. This point shows the exact location of  $ \rho $. Then, with the help of eq.(\ref{eq2}), the equation of the line in the image is obtained. 
Not all detected lines correspond to the desired diagonal patterns. To filter out false positives, constraints on angle and length are applied. Diagonal lines are determined to have angles between 15° and 80° and lengths exceeding the image width based on geometric analysis. In addition, according to Fig.\ref{fi4}, the desired diagonal line is a chord of a right-angled triangle whose perpendicular side is equal to the width of the image, so according to eq.(\ref{eq3}), the length of the chord is greater than each of the sides of the triangle.
%
%
\begin{equation}
\label{eq3}
c=\sqrt{a^2+b^2} \Longrightarrow c>a,b
\end{equation}
\indent
If silence frames are placed at the end of the query, the diagonal line will not continue toward the end of the image, so the length of the line may be smaller than the image-width. To account for that, a 50-pixel margin is introduced for the length of the detected line. The Hough transform algorithm used is Alg.\ref{al2}.\\
\scalebox{0.76}{%
	\begin{algorithm}[H]
		\caption{The proposed Hough transform algorithm}
		\label{al2}
		\textbf{Initialize accumulator H to all zeros\\
			Put image's width =$ l $ }\\
		\For{each edge point $ (x,y) $ in the image}{\For{$ 0 \le \theta \le 180 $}{$ \rho = x\cos\theta + y\sin\theta $\\
				$ H(\theta , \rho) = H(\theta , \rho)+1 $}}
		\textbf{Find the values of $ (\theta , \rho) $ where $ H(\theta , \rho) $ is a local maximum}\\
		\textbf{The detected line $ L(x,y) $ in the image is given by $ \rho = x\cos\theta + y\sin\theta $}\\
		\If{length$ (L(x,y)) > l-50$ \textbf{AND} $15^\circ <\arctan \frac{y}{x}<80^\circ$}{$ L(x,y) $ is goal line}
\end{algorithm}}
\vspace{-2mm}
\section{Experimental Setup and Results}\vspace{-3mm}
\label{se3}
\subsection{Datasets}\vspace{-2mm}
Three datasets were constructed for our study for which the details of their characteristics  are shown in Table \ref{tab1}. \\
\textbf{Farsi1}: Comprises 5348 queries in 0 and 1 classes and 208 reference files, extracted from the big and small Farsdat corpora. Data exhibits variations in speaker demographics (e.g., age, accent, gender), recording conditions, and environmental noise.\\
\textbf{Farsi2}: Contains 300 query-reference pairs from the big Farsdat corpus with matched speakers. The average query length in both the Farsi1 and Farsi2 datasets is 1.8 seconds, while the average reference length is 16.2 seconds.\\
\textbf{Multilingual}: Includes 300 query-reference pairs from the gos-kdl, SWS2013, and NCHLT corpora, focusing on open-environment recordings and matched speakers. The speaker in the query and the associated reference are the same. The average query duration is 2 seconds while the reference is about 9.6 seconds.\vspace{-8mm}
%
%
%
%
\renewcommand{\arraystretch}{1.4}
\begin{table}[!ht]
	\centering
	\caption{\small Characteristics of the evaluation dataset. The parentheses indicate the total length of the sound in 'hours: minutes: seconds' Identical and 'overlap/same' shows whether the query and reference have a common 'speaker/recording' conditions or not.}
	\label{tab1}
	\scalebox{0.68}{%
		\begin{tabular}{|c|p{2.5cm}|c|c|c c|}
			\hline
			\multirow{2}{*}{\textbf{Dataset}} &
			\multirow{2}{*}{\textbf{Language}} &
			\multirow{2}{*}{\textbf{Number of queries (Duration)}} &
			\multirow{2}{*}{\textbf{Number of test (Duration)}} &
			\multicolumn{2}{c|}{\textbf{Query vs. Test items}} \\ \cline{5-6} 
			&                                                  &                &               & \multicolumn{1}{c|}{\textbf{Speakers}} & \textbf{Recordings} \\ \hline
			\large Farsi1       & \large Persian                                          &\large  2674 (4:16:15) &\large  208 (0:53:54) & \multicolumn{1}{c|}{\large overlap}           &\large  same                \\ \hline
			\large Farsi2       & \large Persian                                          &\large  151 (0:6:3)    &\large  26 (0:7:44)   & \multicolumn{1}{c|}{\large same}              &\large  same                \\ \hline
			\large Multilingual &\large  Gronings, African, IsiNdebele(nbl), Sepedi(nso) &\large  152 (0:5:5)    &\large  30 (0:4:48)   & \multicolumn{1}{c|}{\large overlap}           &\large  overlap             \\ \hline
	\end{tabular}}
\vspace{-6mm}
\end{table}
\subsection{Baseline Models}\vspace{-2mm}
Two baseline models by San et al. and Ram et al. \cite{11,12}, were employed for comparison.  San et al. proposed a QbE-STD system with representations derived from a pre-trained self-supervised wav2vec2.0 model (the English monolingual model, wav2vec2-large-960h) and achieved significant performance gains (56–86\% relative) on Australian Aboriginal languages compared to state-of-the-art methods. Their approach incorporated Euclidean similarity and segmental Dynamic Time Warping (DTW).

Ram et al. reformulated the QbE-STD task as a binary image classification problem. Their approach first applies a speech activity detector (VAD) to remove noisy frames, then computes a frame-level similarity matrix between the spoken query and test utterance using cosine similarity of posteriorgram features. The resulting distance matrix visualizations are used to train a CNN classifier. Evaluated on the SWS2013 and QUESST2014 datasets, this model achieved a $10\%$ improvement over baseline methods, demonstrating the efficacy of the CNN-based image classification framework for QbE-STD. This work highlights the value of leveraging computer vision techniques to address speech processing challenges, representing a novel integration of diverse signal processing paradigms.
%
\vspace{-2mm}
\subsection{Evaluation Metric}\vspace{-1mm}
We use the maximum term weighted value (MTWV) \cite{41} as the primary metric. Using the NIST STDEval\footnote{https://www.nist.gov/itl/iad/mig/tools} tool, we calculate MTWVs with the suggested NIST costs (false positive: 1, false-negative: 10) and an empirical prior of 0.0278. An MTWV of 0.48 indicates that the system correctly detects 48\% of all queries being searched, while producing at most 10 false positives for each true positive correctly retrieved. The MTWV ranges from 0 (indicating a system that returns nothing) to 1 (indicating a perfect system detecting all relevant instances with no false positives). In addition, by drawing a confusion matrix, we compare the performance of models \cite{11}.\vspace{-2mm}
\subsection{Experiments}\vspace{-1mm}
To pursue the ablation study evaluations, we compare the performance of each proposed module with existing methods, and finally compare the end-to-end  model with two baseline models. The adjustment parameters of the Hough transform are listed in Table\ref{tab2}. As a pre-processing, all audio data from various sources were standardized to mono 16-bit PCM at 16 kHz (required by XLSR-53).\vspace{-6mm}
\renewcommand{\arraystretch}{1}
\begin{table}[!h]
	\centering
	\caption{Hough transform's parameters}
	\label{tab2}
	\begin{center}
		\scalebox{1}{%
			\begin{tabular}{ |c|c|c|c| }
				\hline
				\multicolumn{2}{|c|}{\textbf{Param. Canny edge detection}} \\
				\hline \hline
				T\_lower & 80 \\ \hline
				T\_upper & 120 \\ \hline \hline
				\multicolumn{2}{ |c| }{\textbf{Param. Hough transform}} \\
				\hline
				\hline
				rho & 1 \\
				\hline
				theta & $ \frac{\pi}{180} $  \\ \hline
				threshold & 30 \\ \hline
				maxLineGap &200 \\ \hline
		\end{tabular}}\vspace{-6mm}
	\end{center}
\end{table}
\subsubsection{Feature Extraction Block Performance}
\begin{figure}[!htb]
	\begin{center}
		{\includegraphics[width=5.3cm, height=3.9cm]{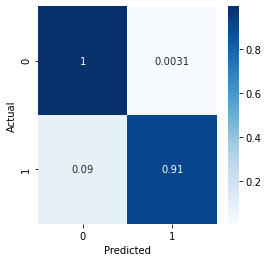}}
		\qquad
		{\includegraphics[width=5.3cm,height=3.9cm]{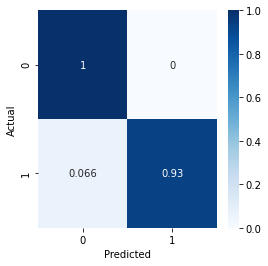}}
		\caption{\small Right side, confusion matrix of model performance in Farsi1 dataset. Left, confusion matrix of model performance in multilingual dataset.}
		\label{fig5}
	\end{center} \vspace{-4mm}
\end{figure}

\begin{figure}[!ht]
	\centering
	\includegraphics[width=5.9cm, height=3.4cm]{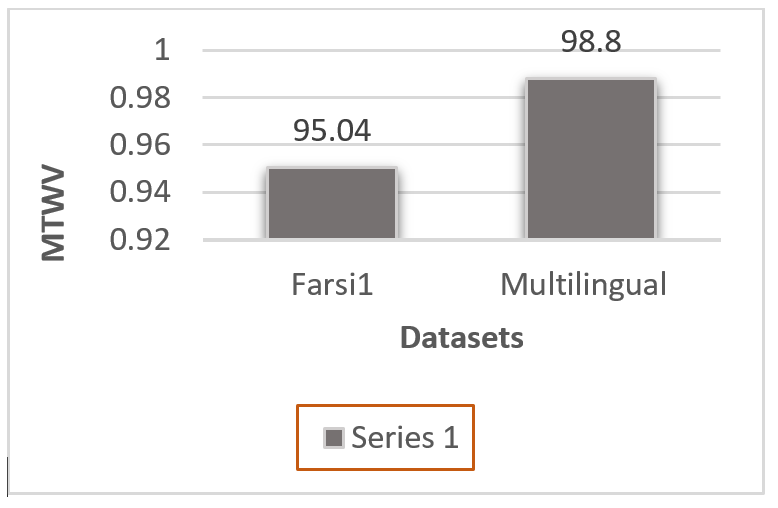}
	\caption{\small Results of MTWV feature extraction model in Farsi1 and Multilingual datasets}
	\vspace{-4mm}
	\label{fi6}
\end{figure}

We posit that the proposed feature extractor module learns language-agnostic, speaker-invariant acoustic representations. To validate this, we evaluate performance across Farsi1 and Multilingual datasets spanning acoustic conditions, languages, and speakers. As shown in Fig. \ref{fig5}, the model demonstrates consistent $100\%$ accuracy in detecting query non-occurrences, with comparable performance in identifying occurrences across datasets. These results corroborate the feature extractor's ability to learn generalizable acoustics that transcend language, speaker, and environmental variations. The model's robust cross-dataset generalization underscores its potential for real-world, multilingual spoken term detection, where adaptability to diverse conditions is crucial. This work highlights the value of developing acoustic features that can be applied broadly, rather than relying on language-specific or speaker-dependent representations.
\\
\indent
The model's MTWV performance on Farsi1 and Multilingual datasets, shown in Fig. \ref{fi6}, demonstrates its language-independent and speaker-invariant capabilities, as well as robust handling of acoustic variations. Notably, the MTWV values approaching 1 underscore the model's exceptional ability to effectively detect query term occurrences, even in challenging conditions. These results highlight the model's potential for real-world, multilingual spoken term detection, where the capacity to generalize across diverse environments is crucial. The consistent high performance showcases the value of the learned acoustic representations in transcending language, speaker, and environmental dependencies.\vspace{-2mm}
\vspace{-1cm}
\subsubsection{Distance Matrix Block Performance}
\begin{figure}[!h]
	\centering
	\subfloat{\includegraphics[width=3.75cm,height=2.9cm]{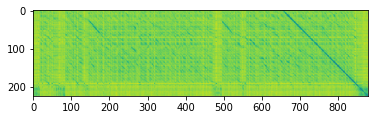}}
	\label{fi7}
	\quad
	\subfloat{\includegraphics[width=3.75cm,height=2.9cm]{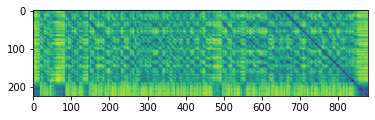}}
	\label{fi8}
	\quad
	\subfloat{\includegraphics[width=3.75cm,height=2.9cm]{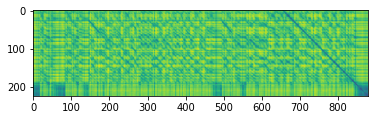}}
	\label{fi9}
	\caption{\small Generated images through the Canberra similarity (\textbf{left}), Cosine similarity (\textbf{middle}), and Euclidean similarity (\textbf{right}) metrics on Farsi2 dataset} 
	\vspace{-6mm}
	\label{fig:fig7}
\end{figure}

The distance matrix calculation module significantly enhances search engine performance by effectively representing common frames. A key challenge is to clearly differentiate between diagonal and non-diagonal modes, which can be addressed by selecting an appropriate distance metric. This metric must accurately distinguish between shared and non-shared pixels, demonstrating a strong capacity for detecting dissimilarities.
In this study, we evaluate the effects of Euclidean, Cosine, and Canberra similarity metrics using the Farsi2 dataset. The resulting distance matrix images for similar classes, generated with these metrics, are presented in Figure. \ref{fig:fig7}.\vspace{-4mm}

%
\indent
The images generated using the Canberra similarity metric exhibit greater transparency compared to those produced by the other metrics, significantly enhancing the quality of class 0 images. The diagonal mode is clearly visible with the Canberra metric, while it is challenging to distinguish in the Euclidean and Cosine metrics. The result of the experiment on Farsi2 data is as follows:
\begin{figure}[!ht]
	\centering
	\subfloat{\includegraphics[width=3.6cm,height=3cm]{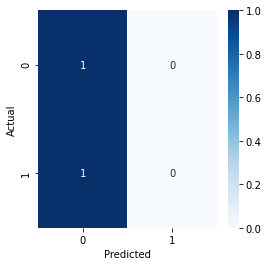}}
	\label{fi11}
	\quad
	\subfloat{\includegraphics[width=3.6cm,height=3cm]{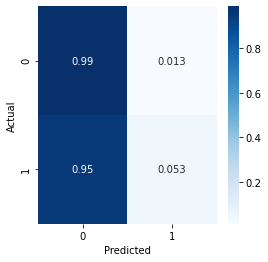}}
	\label{fi12}
	\quad
	\subfloat{\includegraphics[width=3.6cm,height=3cm]{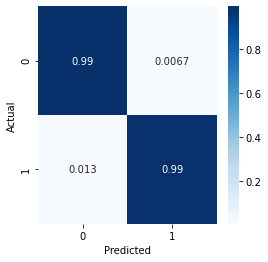}}
	\label{fi10}
	\caption{\small The proposed method confusion matrix using three metrics (from right to left) Canberra, Euclidean and Cosine similarity on Farsi2 dataset}
	\vspace{-4mm}
	\label{fi13}
\end{figure}
The Canberra similarity metric demonstrates superior performance compared to the Cosine and Euclidean metrics due to its consideration of point-to-point similarity, accounting for the size of values and suitability for high-dimensional, continuous, and variable data. This metric has improved by $49\%$ over the Cosine similarity metric and $47\%$ over the Euclidean similarity metric. Figure \ref{fi14} depicts the MTWV (Maximum Term-Weighted Value) of the model using these metrics. As illustrated, the Canberra similarity criterion has achieved the highest MTWV value, further emphasizing its effectiveness in this context.\vspace{-4mm}
\begin{figure}[!htb]
	\centering
	\includegraphics[width=6.6cm, height=3.9cm]{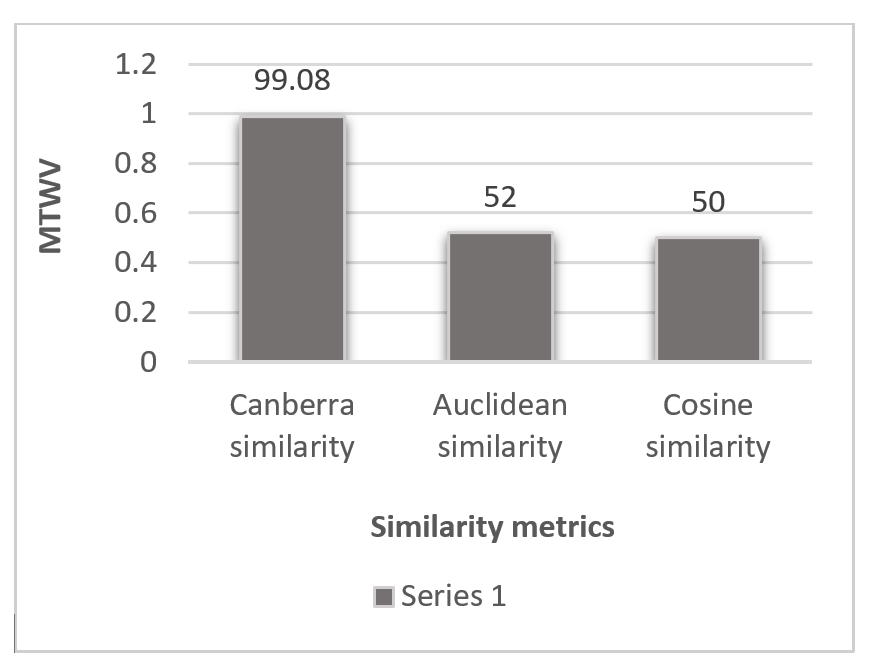}
	\caption{\small MTWV of the proposed method using  Canberra, Euclidean, and Cosine similarity metrics via Farsi2 dataset.}\vspace{-6mm}
	\label{fi14}
\end{figure}
\vspace{-2mm}
\subsubsection{Template Recognition Block Performance}
\begin{figure}[!ht]
	\centering
	\subfloat{\includegraphics[width=5.7cm,height=4cm]{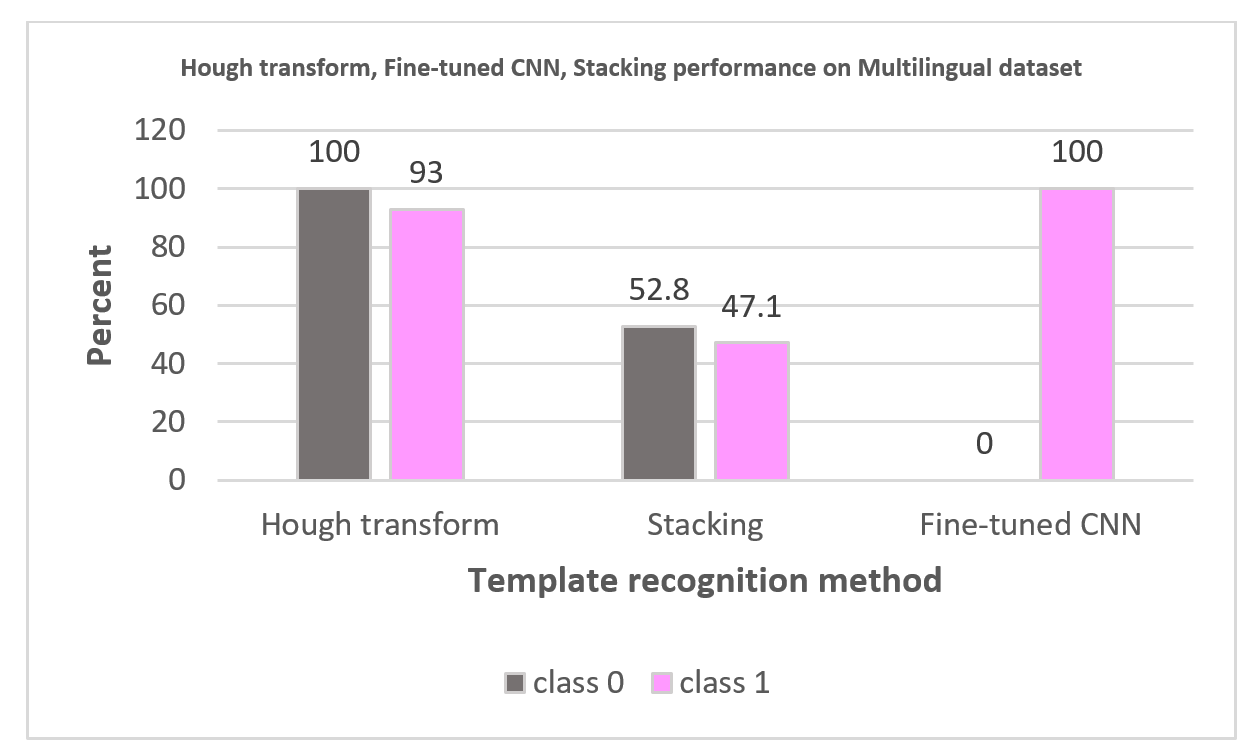}}
	\label{fi15}
	\quad
	\subfloat{\includegraphics[width=5.7cm,height=4cm]{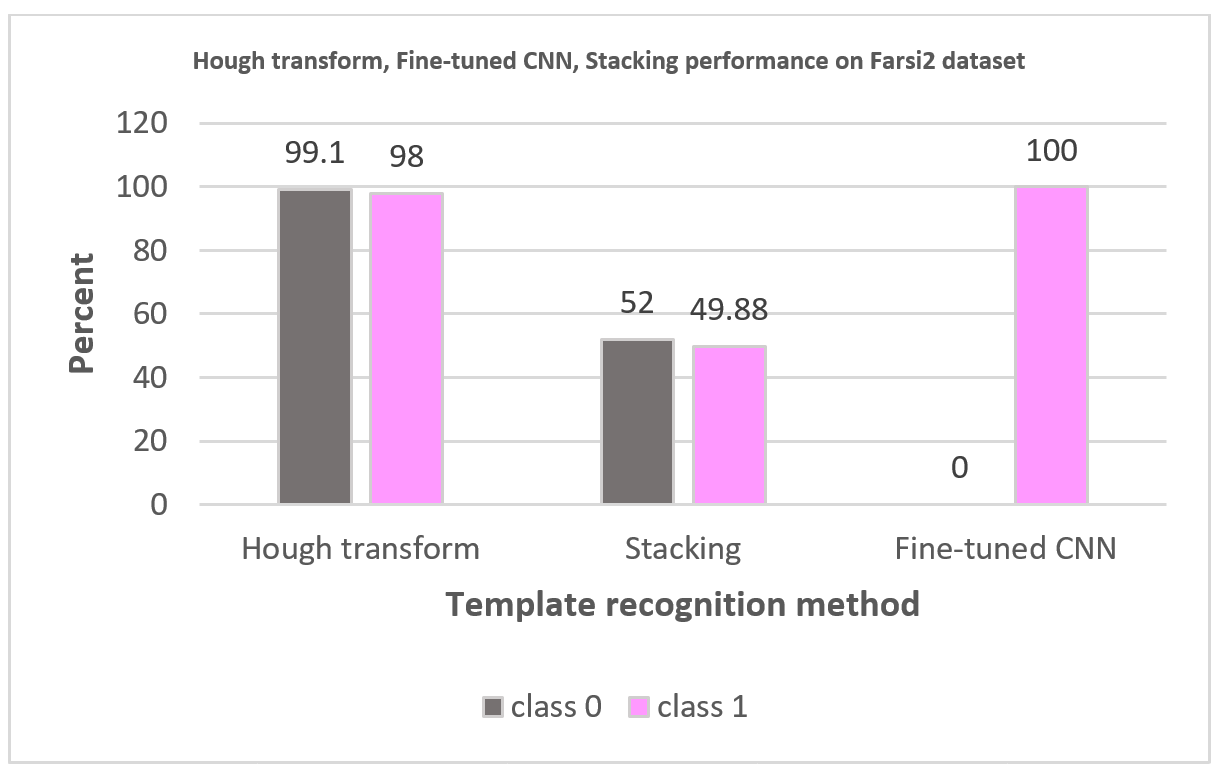}}
	\label{fi16}
	\caption{\small Classification accuracy of template recognition models (Hough transform, Fine-tuned CNN, and Stacking) in each class on Farsi2 and Multilingual datasets.} \vspace{-8mm}
	\label{fi17}
\end{figure}

In this experiment, we design a classifier to separate diagonal and non-diagonal images using CNNs and a stacking ensemble method. The performance of these approaches is then compared to the Hough transform. For the CNN model, we utilize the architecture presented in \cite{12}. The stacking ensemble combines 5 base classifiers: Decision Tree, K-nearest neighbors, Logistic Regression, Gaussian Naive Bayes, and SVM.
The experiment is conducted using the Farsi1 dataset, which is split into training, validation, and testing sets. The Farsi2 and Multilingual datasets are used for testing. The training set consists of 1650 samples with label 1 and 1656 samples with label 0. The validation set includes 602 samples for each class 0 and 1.
\\
\indent
\textbf{CNN-based: }
In this experiment, we employ the pre-trained weights of the CNN model proposed by Ram et al. \cite{12} as a starting point. By adding a fully connected layer, we fine-tune the network on our dataset. CrossEntropyLoss, Softmax classifier and Adam optimizer with a learning rate of 0.001, which is reduced by a factor of 0.1 in every seven epochs is used for training of the model, and distance matrix images with a fixed size of $ 750\times200 $ as inputs.
Figure \ref{fi17} presents the performance of the fine-tuned CNN model separately for each class. As depicted, the model exhibits poor performance. This could be attributed to the weakness of the pre-trained model or the changes in the input image size.
The size of the distance matrix images is proportional to the number of query and reference frames. Since the audio files in the Farsi1 dataset have varying durations, the produced images also differ in size. Neural networks require inputs with a fixed size. Changing the size of the distance matrix images leads to the loss of time information and visual information, particularly affecting the diagonal pixels.
This problem becomes more pronounced in real-world scenarios when the reference files have longer durations, potentially causing significant issues.

\textbf{Stacking-based: } In this experiment, we evaluated the performance of five before-mentioned classifiers using our dataset. Logistic Regression and SVM yielded the best results, leading to their selection for further analysis. We constructed a stacking model utilizing Logistic Regression as the meta-model. To enhance the model's performance, we applied Principal Component Analysis (PCA) for dimensionality reduction, reducing the feature dimensions to one. The input data for this model consists of images sized $ 300\times300 $ pixels.
As shown in Figure \ref{fi17}, the stacking-based method outperforms the fine-tuned CNN model. This can be attributed to the stacking model's ability to achieve better results with fewer data and a simpler algorithm compared to the CNN approach. Figure \ref{fi17} clearly demonstrates the superior accuracy of the proposed stacking model compared to the other tested methods. On average, the Hough transform exhibits a $47\%$ improvement over the CNN and stacking models.
\vspace{-5mm}

\subsubsection{QbE-STD Performance}
Finally, we evaluated the proposed model as an integrated system and compared it against two baseline models, San et al. \cite{11} and Ram et al. \cite{12}. The experiment utilizes the Farsi2 and Multilingual datasets, with results summarized in Table \ref{tab3}. The advantages of the proposed method are outlined as follows:
\begin{enumerate}
	\item \textbf{Consistent Performance Across Languages}: The proposed method performs uniformly across different languages, achieving a $19\%$ improvement over Ram et al.'s method on the Multilingual dataset and a $54\%$ improvement on the Farsi2 dataset.
	\item \textbf{No Need for Training or Labels}: Unlike Ram et al.'s approach, the proposed method does not require training or labeled data.
	\item \textbf{Robustness to Dataset Variability}: Ram et al.'s model, which was trained on the SWS2013 dataset, performs better on the Multilingual dataset than on Farsi2. In contrast, the proposed method leverages the pre-trained XLSR-53 model, trained on 53 languages, ensuring good performance across a wider range of languages.
	\item \textbf{Preservation of Time Information}: The method proposed by Ram et al. requires fixed-size distance matrix images, which can lead to the loss of temporal information. The proposed method does not necessitate resizing, facilitating accurate query localization within reference files.
	\item \textbf{Close to State-of-the-Art Performance}: A comparison with San et al.'s method \cite{11} reveals that the proposed method achieves results comparable to state-of-the-art techniques in QbE-STD.
	\item \textbf{Line Recognition Capability}: The proposed method, based on the principles of the Hough transform, can identify multiple lines in an image, allowing it to recognize and count repeated queries in the reference. In contrast, the DTW method identifies only a single minimal path \cite{17}.
	\item \textbf{Faster Query Processing}: The proposed method can process a query of approximately 4 seconds in 26 Farsi2 reference files, operating 5\% faster than the method presented by Sun et al. \cite{11}.
	\vspace{-10mm}	
\end{enumerate}

\renewcommand{\arraystretch}{1}
\begin{table}[!htb]
	\centering
	\caption{Comparison of the proposed method with baseline methods}
	\label{tab3}
	\scalebox{1}{%
		\begin{tabular}{|c|c|c|c|}
			\hline
			\textbf{Method}	& Ram et al. \cite{12} & San et al. \cite{11} & Our proposed \\ \hline
			\textbf{Farsi2 (MTWV)}	& 0.45 & 1 & 0.9925 \\ \hline
			\textbf{Multilingual (MTWV)}	& 0.69 & 1 & 0.988 \\ \hline
			\textbf{Time (seconds)}	& 20 & 84 & 79 \\ \hline
	\end{tabular}}
\end{table}

\vspace{-1cm}
\section{Conclusions}
\vspace{-2mm}
\label{se4}
This study proposes a novel language-independent transformer-based approach for Query-by-Example Spoken Term Detection to address the increasing demand for efficient audio retrieval systems. The method provides a discriminative detection framework that outperforms Dynamic Time Warping and CNN-based systems in terms of time efficiency and accuracy. Key advantages of the proposed approach include language-independent performance, the ability to distinguish between query occurrence and non-occurrence, and significant improvements over baseline models.  This study represents a significant advancement in QbE-STD, paving the way for more robust and versatile voice search and audio data management solutions.

\vspace{-2mm}

%
%
%
 \bibliographystyle{splncs04}
 \bibliography{mybibfile}

\end{document}